\documentclass{article}

\PassOptionsToPackage{numbers, compress}{natbib}

\usepackage[final]{neurips_2019}

\usepackage[utf8]{inputenc} 
\usepackage[T1]{fontenc}    
\usepackage{hyperref}       
\usepackage{url}            
\usepackage{booktabs}       
\usepackage{amsfonts}       
\usepackage{nicefrac}       
\usepackage{microtype}      
\usepackage[pdftex]{graphicx} 
\usepackage{caption}
\usepackage{subcaption}

\usepackage{kotex}          
\usepackage{algorithm}
\usepackage{algorithmic}
\usepackage{xcolor}
\usepackage{amssymb, commath, bm, amsbsy}
\usepackage{multirow}
\usepackage{amsmath}
\usepackage[normalem]{ulem}

\title{Domain-Agnostic Few-Shot Classification \\ by Learning Disparate Modulators}

\vspace{-0.5cm}
\author{
	Yongseok Choi \hspace*{7pt} Junyoung Park \hspace*{7pt} Subin Yi \hspace*{7pt} Dong-Yeon Cho \\ [0.2cm]
	T-Brain, AI Center, SK Telecom\\
	\{yschoi, jypark, yisubin, dycho24\}@sktbrain.com
}

\begin{document}

\maketitle





\vspace{-0.6cm}
\begin{abstract}
\vspace{-0.2cm}

Although few-shot learning research has advanced rapidly with the help of meta-learning, its practical usefulness is still limited because most of them assumed that all meta-training and meta-testing examples came from a single domain. 
We propose a simple but effective way for few-shot classification in which a task distribution spans multiple domains including ones never seen during meta-training.
The key idea is to build a pool of models to cover this wide task distribution and learn to select the best one for a particular task through cross-domain meta-learning.
All models in the pool share a base network while each model has a separate modulator to refine the base network in its own way.
This framework allows the pool to have representational diversity without losing beneficial domain-invariant features.
We verify the effectiveness of the proposed algorithm through experiments on various datasets across diverse domains.

\end{abstract}
\vspace{-0.5cm}
\section{Introduction}
\vspace{-0.2cm}

Few-shot learning in the perspective of meta-learning aims to train models which can quickly solve novel tasks or adapt to new environments with limited number of examples. 
In case of few-shot classification, models are usually evaluated on a held-out dataset which does not have any common class with the training dataset.
In the real world, however, we often face harder problems in which novel tasks arise arbitrarily from many different domains even including previously unseen ones.

In this study, we propose a more practical few-shot classification algorithm to generalize across domains beyond the common within-domain setup.
Our approach to cover a complex task distribution is to construct a pool of multiple models and learn to select the best one given a novel task. 
This recasts task-specific adaption across domains as a simple selection problem, which makes learning more effective.
Furthermore, by enforcing all models to share some of parameters, each model could keep important domain-invariant features while the pool maintains diversity as a whole.

Experimental results show that our algorithm could perform few-shot classification tasks from multiple meta-trained domains without having any domain identifiers at meta-test time.
They also reveal that the proposed algorithm could be applied successfully to novel tasks from unseen domains.


\vspace{-0.2cm}
\section{Methods}
\vspace{-0.2cm}

\subsection{Problem statement}
\vspace{-0.2cm}

Our objective is to build a domain-agnostic meta-learner beyond the common meta-learning assumptions, i.e. meta-training within a single domain and meta-testing within the same domain, presuming that one domain corresponds to one dataset. 
After meta-training over diverse datasets, we apply the trained meta-learner to two types of more general few-shot classification tasks which require the inter-domain generalization capability of the meta-learner.

One is few-shot classification tasks sampled from held-out classes of the datasets used during the meta-training time without knowing from which dataset each task is sampled. This could be used to evaluate whether the meta-learner is capable to adapt to a complex task distribution across multiple domains.
We also tackle few-shot classification tasks sampled from datasets never seen during the meta-training, which requires generalization to out-of-distribution tasks in domain-level.

\vspace{-0.2cm}
\subsection{Building a pool of embedding models from diverse source domains}
\vspace{-0.2cm}

\begin{figure}
    \centering
    \includegraphics[width=0.95\linewidth]{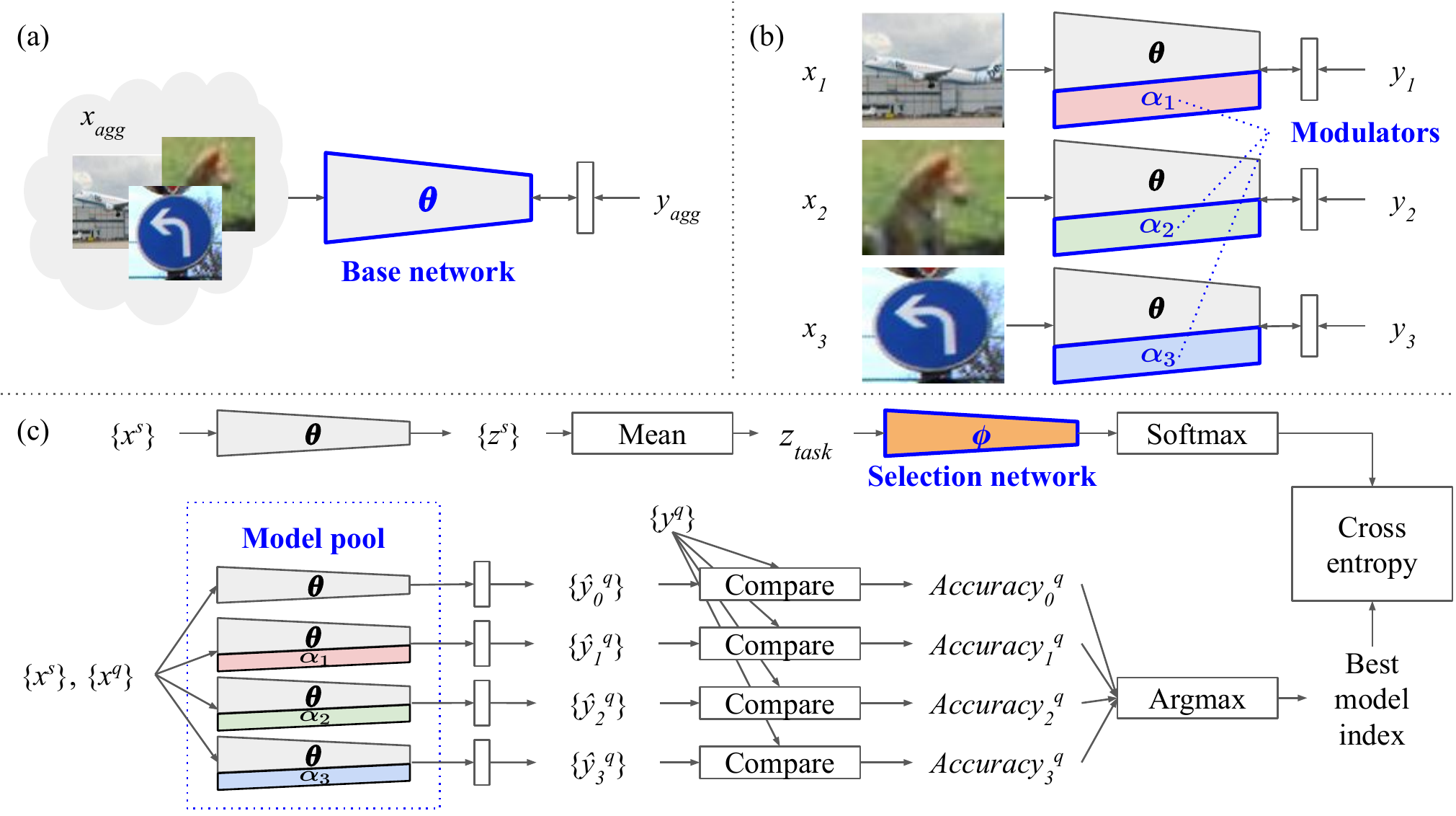}
    \vspace{-0.2cm}
    \caption{
    Training (a) a base network ($\theta$), (b) modulators ($\{\alpha_i\}$), (c) a selection network ($\phi$). ($x^n_m$: an example, $y^n_m$: a label, $\hat{y}^n_m$: a prediction, $z^n_m$: an embedding vector of a support ($n$=$s$) or query ($n$=$q$) set in a domain $m$.)
    }
    \vspace{-0.2cm}
    \label{fig:training}
\end{figure}

Basically, we learn an embedding model in a similar way to Prototypical Network (ProtoNet) \cite{DBLP:conf/nips/SnellSZ17_protoNet} over various datatsets, each of which is considered to define its own domain.
Few-shot classification can be performed across various domains by comparing distances to class prototypes from a query embedding vector on a metric space defined by the learned embedding model.
However, since it is hard to map a complex task distribution spanning diverse domains into a single space, we build a pool of embedding models each of which has its own metric space.
For inference, one model can be chosen as the best fit for a particular task, or we can infer by averaging outputs from all models.

Rather than training an individual model separately, we first train a base network shared by all models over an aggregation of all source datasets following the supervised learning procedure (Figure \ref{fig:training}(a)).
Then, we build one model per source domain by adding a per-model modulator on top of the frozen base network and training each modulator on one dataset by performing metric-based meta-learning in the same way as the ProtoNet (Figure \ref{fig:training}(b)).
The modulators are inserted into the base network in a per-layer basis (Figure \ref{fig:arch} in the supplementary material).
The rationale behind this is to let the pool have diversity, which is desirable to represent a complex task distribution, with minimal parameter overhead by the modulators and capture good domain-invariant features by the base network.






The overall training procedure is summarized in Algorithm \ref{alg:main} in the supplementary material.

\vspace{-0.2cm}
\subsection{Learning to select the best model from the pool for a target task}
\vspace{-0.2cm}

After the construction of the pool, we learn a meta-model which predicts the best single model from the pool for a given task as described in Algorithm \ref{alg:main} in the supplementary material and Figure \ref{fig:training}(c).
By training this model over a number of episodes sampled from all source datasets, we expect this ability to be generalized to novel tasks from various domains including unseen domains.

This meta-model parameterized by $\phi$, called a model selection network, is trained in order to map a task representation $z_{task}$ to an index of the best model in the model pool over the sampled episodes. 
The task representation is obtained by passing all examples in the support set of the task through the base network and averaging all resulting embedding vectors.
The index of the best model, which is the ground truth label for training the selection network, is generated by measuring the classification accuracy of all models in the pool with the query set and picking one which has the highest accuracy. 
This simple classification is much easier to learn than manipulating high-dimensional parameters directly, which makes our approach for task-specific adaptation more effective. 

\vspace{-0.2cm}
\section{Experiments}\label{sec:exp}

\vspace{-0.2cm}
\subsection{Setup}
\vspace{-0.2cm}


\subsubsection{Datasets} 
\vspace{-0.2cm}

We use eight image classification datasets, denoted as Aircraft, CIFAR100, DTD, GTSRB, ImageNet12, Omniglot, SVHN, UCF101 and Flowers, from the Visual Decathlon dataset \cite{DBLP:conf/nips/RebuffiBV17_visualdecathlon} for evaluation.
All datasets are resplit for the purpose of few-shot classification.
More detailed information about the datatsets can be found in Section \ref{sec:datasets} in the supplementary material.
%

\vspace{-0.2cm}
\subsubsection{Algorithms}
\vspace{-0.2cm}


We denote inference methods using the model picked by the selection network as \textit{DoS} (Domain-generalized method by Selection) and \textit{DoS-Ch}, which modulates the base network with convolution 1$\times$1 and chnannel-wise modulation (i.e. scaling and biasing) respectively.
We also explore inference methods, \textit{DoA} (Domain-generalized method by Averaging) and \textit{DoA-Ch}, to generate an output by averaging class prediction probabilities of all constituent models modulated by the above-mentioned two types of modulators. 
Two modulation schemes are explained further with their parameter overhead in Section \ref{sec:architectures} in the supplementary material.


Our algorithms are compared with \textit{Fine-tune}, \textit{Simple-Avg}, \textit{ProtoNet} \cite{DBLP:conf/nips/SnellSZ17_protoNet}, \textit{FEAT} \cite{DBLP:journals/corr/abs-1812-03664_feat} and \textit{ProtoMAML} \cite{DBLP:journals/corr/abs-1903-03096_metaDataset}.
\textit{Fine-tune} is a baseline method to add a linear classifier on top of the pre-trained base network and fine-tune it with the support set examples for 100 epochs during meta-testing.
In \textit{Simple-Avg}, we train an embedding model independently on each source domain without sharing any parameters with others in the same way as \textit{ProtoNet} and perform inference by averaging class prediction probabilities of all these models. 
\textit{FEAT} and \textit{ProtoMAML} are the state-of-the-art algorithms focusing on single domain and cross-domain setups respectively.
\textit{TADAM} \cite{DBLP:conf/nips/OreshkinLL18} was also tested but excluded from the results because its training did not converge in our multi-domain experiments.

All results are produced by our own implementations including `further adaptation' introduced in \cite{chen2018a_closerLook}.
Other details about the training are explained in Section \ref{sec:training_details} in the supplementary material.

\vspace{-0.2cm}
\subsection{Results}
\vspace{-0.2cm}
We compare our methods with other algorithms in 5-way 5-shot setting on both seen and unseen domains.
More results including 5-way 1-shot cases are in Section \ref{sec:add_exp_results} in the supplementary material. 

\begin{figure*}[t!]
	\begin{subfigure}{\textwidth}
    	\centering
        \includegraphics[width=0.98\linewidth]{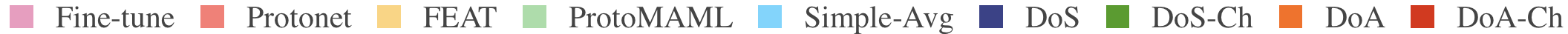}
		\vspace{0.1in}
    \end{subfigure}
    \begin{subfigure}{0.48\textwidth}
    	\centering
        \includegraphics[width=\linewidth]{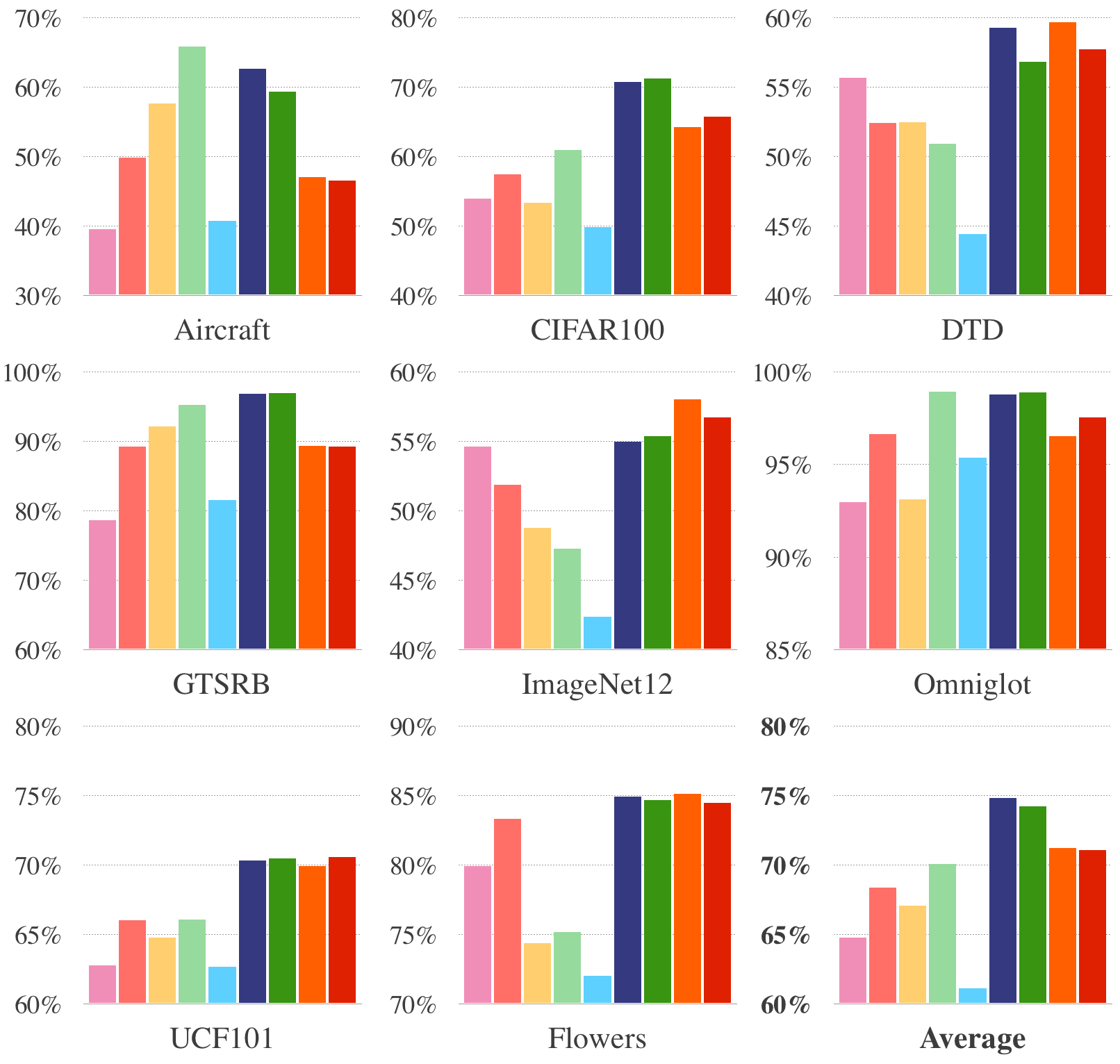}
		\caption{On seen domains.}
    \end{subfigure}
    \hfill
    \begin{subfigure}{0.48\textwidth}
        \centering
		\includegraphics[width=\linewidth]{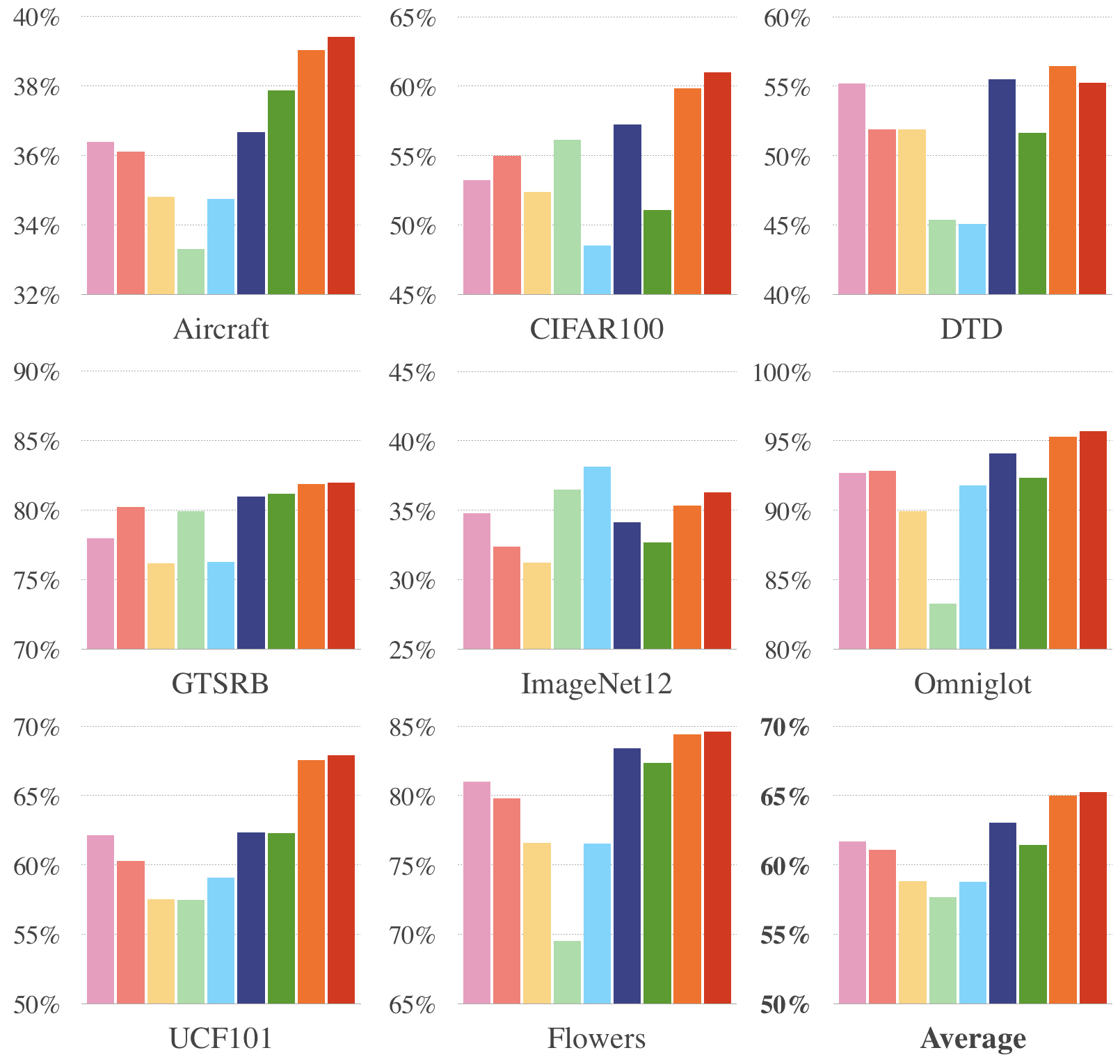}
		\caption{On unseen domains.}
    \end{subfigure}
    \caption{5-way 5-shot test accuracy on various datasets.
    }
    \vspace{-0.2cm}
    \label{figure:results_5w5s}
    \vspace{-0.2cm}
\end{figure*}

\vspace{-0.2cm}
\subsubsection{Few-shot classification on seen domains}
\vspace{-0.2cm}

Figure \ref{figure:results_5w5s}(a) shows the test accuracy values on various seen domains.
We evaluate the above-mentioned algorithms in a multi-domain test setup constructed using all available datasets. 
Specifically, we meta-train a model for each algorithm on all available eight datasets.
Then, we meta-test the trained model for various tasks sampled from these eight datasets without knowing which dataset the task comes from
Our methods, \textit{DoS}, \textit{DoS-Ch}, \textit{DoA} and \textit{DoA-Ch}, outperform other few-shot classification methods in most cases.
Although \textit{FEAT} and \textit{ProtoMAML} adopt their own task-specific adaptation schemes, they do not seem as effective as ours under this complex task distribution across various domains.
\textit{ProtoMAML} shows comparable results in some cases, but much inferior results in other cases.
\textit{ProtoNet} seems relatively stable, but does not produce better results than ours either.

Our selection-based methods, \textit{DoS} and \textit{DoS-Ch}, perform better than our averaging methods on seen domains.
This implies that the learned selection network is working properly, which is highly likely to select the model with the modulator trained on the same domain as the given task even without any information about the domain at testing time.
Another implication is that the best single model might be better than the averaging approach if the model from the same domain exists.
It is also worth noting that \textit{DoS-Ch} is quite competitive despite less number of additional parameters than \textit{DoS}.


\vspace{-0.2cm}
\subsubsection{Few-shot classification on unseen domains}
\vspace{-0.2cm}

We also report the test results on unseen domains in Figure \ref{figure:results_5w5s}(b).
Given a target dataset for test, we train all models using on seven other datasets.  
Our methods still outperform other algorithms in this more challenging setting, which reveals that our methods can be generalized to novel domains as well.
Differently from the seen domain cases, our averaging methods, \textit{DoA} and \textit{DoA-Ch}, perform better than all other algorithms including our selection-based methods.
It seems to make sense since the averaging could induce naturally synergy between beneficial models due to their similar output patterns even if we do not know which models are beneficial to a given task at testing time.
However, our averaging methods outperform significantly \textit{Simple-Avg}, which implies that our unique architecture to encourage keeping domain-invariant features, i.e. embedding models with parameter sharing, is another key factor to the high performance of the averaging methods.

\vspace{-0.2cm}
\section{Related works}
\vspace{-0.2cm}

Meta-learning is one of the most popular techniques to solve the few-shot learning problems, which includes learning a task-invariant metric space \cite{DBLP:conf/nips/SnellSZ17_protoNet,DBLP:conf/nips/VinyalsBLKW16}, learning to optimize \cite{DBLP:conf/nips/AndrychowiczDCH16,DBLP:conf/iclr/RaviL17} or learning weight initialization \cite{DBLP:conf/icml/FinnAL17_MAML,DBLP:journals/corr/abs-1803-02999}. 
Follow-up studies showed that metric-based meta-learning could be improved further by learning task-specific adaptation on the learned space \cite{DBLP:conf/cvpr/GidarisK18,DBLP:conf/nips/OreshkinLL18,DBLP:conf/cvpr/QiaoLSY18,rusu2018metalearning,DBLP:journals/corr/abs-1812-03664_feat}. 

Recent few-shot learning studies have tried to tackle challenging problems under more realistic assumptions. 
Some studies dealt with few-shot learning under domain shift between training and testing \cite{DBLP:conf/uai/KangF18}\cite{DBLP:conf/eccv/WangH16}.
More realistic benchmark was proposed for few-shot learning to overcome limitations of the current popular benchmarks including the lack of domain divergence \cite{DBLP:journals/corr/abs-1903-03096_metaDataset}. 
Similar to our approach, a few suggestions combined multiple models to benefit from their diversity \cite{DBLP:journals/corr/abs-1903-11341,DBLP:journals/corr/abs-1904-08479,DBLP:journals/corr/abs-1904-05658}.
However, they considered an ensemble with independent models unlike ours with the shared network.
Our network architecture is inspired by the parameter sharing strategies for multi-task learning  \cite{DBLP:journals/corr/Ruder17a} and multi-domain learning with domain-specific adaptation \cite{DBLP:conf/cvpr/RebuffiBV18} because they have been known to lead to efficient parameterization and positive knowledge transfer between heterogeneous entities.
\vspace{-0.2cm}
\section{Conclusion and future works}
\vspace{-0.2cm}

We propose a new few-shot classification method generalizing to various domains including unseen domains. 
The core idea is to build the pool of embedding models, each of which is diversified by its own modulators while sharing most of parameters with others, and learn to select the best model for a target task through cross-domain meta-learning.
Experiments show that the proposed method outperforms an ensemble of models trained separately on every different dataset and the state-of-the-art few-shot classification algorithms.

We believe that there is still a large room for improvement in this challenging task. 
It would be one promising extension to find the optimal way to build the pool without being confined by the policy of one model per dataset so that it can work even with a single source domain with large diversity.
Soft selection or weighted averaging can be also thought as one of future research directions because a single model or uniform averaging is less likely to be optimal. 
We can also consider a more scalable extension to allow continual expansion of the pool only by training a modulator for an incoming source domain without re-training all existing models in the pool.

\bibliography{references}
\bibliographystyle{abbrv}

\newpage
\clearpage
\appendix
\section*{Supplementary Materials}
\section{Datasets}\label{sec:datasets}

In our experiments, we use the Visual Decathlon dataset \cite{DBLP:conf/nips/RebuffiBV17_visualdecathlon} which consists of ten datasets for image classification listed below.
\begin{itemize}
    \item FGVC-Aircraft Benchmark (Aircraft, A) \cite{DBLP:journals/corr/MajiRKBV13}
    \item CIFAR100 (CIFAR100, C) \cite{Krizhevsky2009LearningML}
    \item Daimler Mono Pedestrian Classification Benchmark (DMPCB)\cite{Munder06}
    \item Describable Texture Dataset (DTD, D) \cite{Cimpoi14}
    \item German Traffic Sign Recognition Benchmark (GTSRB, G) \cite{STALLKAMP2012323}
    \item ImageNet ILSVRC12 (ImageNet12, I) \cite{ImageNet:ILSVRC15}
    \item Omniglot (Omniglot, O) \cite{Lake1332}
    \item Street View House Numbers (SVHN) \cite{Netzer11}
    \item UCF101 (UCF101, U) \cite{DBLP:journals/corr/abs-1212-0402}
    \item Flowers102 (Flowers, F) \cite{Nilsback08}
\end{itemize}
The categories and the number of images of each dataset are significantly different as well as the image size. 
All images have been resized isotropically to $72\times72$ pixels so that each image from various domains has the same size. 

Daimler Mono Pedestrian Classification has only 2 classes, pedestrian and non-pedestrian. 
We excluded it from our experiments as we are considering 5-way classification tasks. 
SVHN was also excluded since SVHN has only 10 digit classes from 0 to 9, which were too few to split for meta-training and meta-testing.
To use the remaining eight datasets for multi-domain few-shot classification, we divide the examples into roughly 70\% training, 15\% validation, and 15\% testing classes. 
For ILSVRC12, we follow the split of Triantafillou {\it et al.} \cite{DBLP:journals/corr/abs-1903-03096_metaDataset} to adopt class hierarchy, and we use random class splits for other datasets. 
The number of classes at each split is shown in Table \ref{table:decathlon}.
We only use train and validation sets of the Visual Decathlon because the labels of the test set is not publicly available. 

\begin{figure}[htbp]
    \centering
    \includegraphics[width=1.0\linewidth]{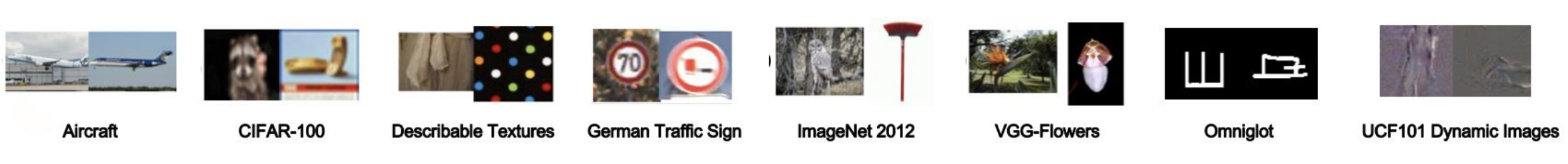}
    \label{fig:datasets}
\end{figure}

\begin{table}[htbp]
  \caption{The details of datasets used in our experiments. 
  }
  \label{table:decathlon}
  \vskip 0.15in
  \begin{center}
  \begin{small}
  \begin{sc}
  \begin{tabular}{ lccccc}
     \toprule
            &&&& Splits \\
            \cmidrule(r){4-6}
     Dataset    & \# data   & \# classes    & Train & Val   & Test\\
     \midrule
     aircraft   & $6667$    & $100$         & $70$  & $15$  & $15$\\
     CIFAR100   & $50000$   & $100$         & $70$  & $15$  & $15$\\
     DTD        & $3760$    & $47$          & $32$  & $7$   & $8$\\
     GTSRB      & $39209$   & $43$          & $30$  & $6$   & $7$\\
     Imagenet12 & $1281167$ & $1000$        & $712$ & $158$ & $130$\\
     Omniglot   & $25968$   & $1623$        & $1136$& $243$ & $244$\\
     UCF101     & $9537$    & $101$         & $70$  & $15$  & $16$\\
     Flowers    & $2040$    & $102$         & $70$  & $16$  & $16$\\
     \bottomrule
  \end{tabular}
  \end{sc}
  \end{small}
  \end{center}
\end{table}
\section{Architectures}\label{sec:architectures}

\begin{figure}
    \centering
    \includegraphics[width=1.0\linewidth]{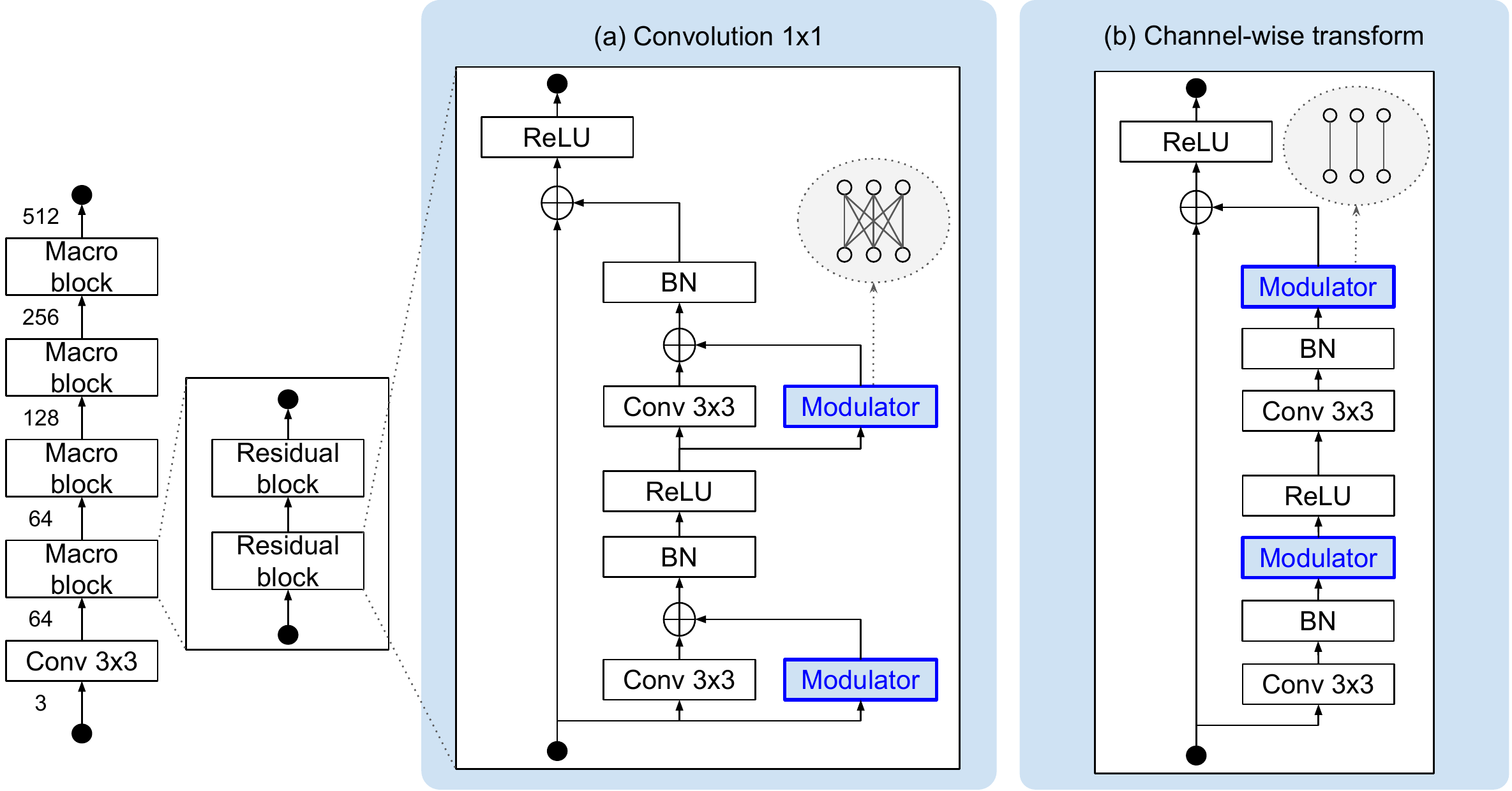}
    \caption{Architecture of the ResNet-18-based embedding network $f_e$.}
    \label{fig:arch}
\end{figure}

Figure \ref{fig:arch} shows the architecture of the embedding network $f_E(\cdot; \theta, \alpha_i)$, which processes an input image and produces a 512-dimensional embedding vector. 
The embedding network is based on the ResNet-18 architecture \citep{DBLP:conf/cvpr/HeZRS16_resNet}, which consists of one convolutional layer with 64 $7\times7$ filters followed by 4 macro blocks, each having 64-128-256-512 $3\times3$ filters.
Figure \ref{fig:arch}(a) and Figure \ref{fig:arch}(b) depict how the base network is modulated by the convolution $1\times1$ modulator and the channel-wise transform modulator, respectively. These modulators are placed within each residual block of the macro blocks, same as the previous works in \citep{DBLP:conf/cvpr/RebuffiBV18} and \citep{DBLP:conf/aaai/PerezSVDC18}.

The number of parameters for two modulators are shown in Table \ref{table:num_params}. 
The values on the first row are the number of modulator parameters that are additionally applied to the embedding network. 
Note that the channel-wise transform modulator has much fewer parameters than the convolution 1$\times$1 modulator. 
In particular, the channel-wise transform modulator has negligible number of parameters compared to that of the base network, i.e., ResNet-18.
For each embedding model, the convolution $1\times1$ modulator has about 10\% of the number of parameters that the base network has whereas the channel-wise transform modulator requires only less than 1\%.

The selection network $f_S(\cdot; \phi)$ is a two-layered MLP (multi-layer perceptron) network, which receives an embedding vector produced by the embedding network as an input and performs the best model index prediction. 
Two layers are a linear layer of $512\times128$ and a linear layer of $128\times(M+1)$, where $M$ is the number of source domains. 

\begin{table*}
    \vskip 0.15in
    \begin{center}
    \begin{sc}
    \begin{tabular}{ ccc}
        \toprule
                & Convolution 1x1 & Channel-wise transform \\
                \midrule
        Modulators $\{\alpha_i\}^8_{i=1}$ & 9,795,584 & 61,440\\
        Base network $\theta$ & 11,176,512 & 11,176,512 \\
        Selection network $\phi$ & $66,696$ & $66,696$ \\
        \midrule
        Sum & $21,038,792$ & $11,304,648$ \\
        \bottomrule
    \end{tabular}
    \end{sc}
    \end{center}
    \caption{The comparison of the number of parameters for convolution $1\times1$ and channel-wise transform modulators. This is the case when the number of source domains is 8. }
    \label{table:num_params}
\end{table*}

\section{Training details}\label{sec:training_details}

Algorithm \ref{alg:main} describes the overall training procedure to construct the model pool and the selection network.
Although we trained three components in a sequential manner, joint training of these components seems to make sense also.

For the fair comparison with \textit{Fine-tune} method, we also apply algorithm-specific refinement at meta-testing time, inspired by `further adaptation' in \cite{chen2018a_closerLook}, to all other algorithms including ours. 
A linear classifier is placed on top of the embedding network of the \textit{ProtoNet}, the self-attention module of the \textit{FEAT} or the modulated embedding network of our models. 
During meta-testing, other parameters are fixed and the classifier is fine-tuned using the support examples for 100 iterations per episode.
In case of \textit{FEAT}, the classifier is trained for 100 epochs per query example not per episode because \textit{FEAT} modulates a representation space for each query.
We also adjust the number of adaptation of the \textit{ProtoMAML} to 100 for the better task-adaptation as done in \cite{chen2018a_closerLook}. 

The hyperparameters including the learning rate are selected by grid search based on the validation accuracy.
For \textit{FEAT} and \textit{ProtoMAML}, Adam optimizer is used for training and the learning rate and weight decay are set to be 0.0001. 
Other models are also trained using Adam optimizer with the learning rate 0.001 but without any regularization including the weight decay. 

All reported test accuracy values are averaged over 600 test episodes with 10 queries per class. 

\begin{algorithm}[tb]
\begin{algorithmic}
\STATE {\bfseries Input:} Training data from $D_S=\{D_{S_i}\}^{M}_{i=1}$, embedding networks $f_e(\cdot)$, a selection network $f_s(\cdot)$ ($D_{S_1}, \cdots, D_{S_M}$: source domains where $M$ is the number of source domains, $D_T$: a target domain)
\STATE {\bfseries Output:} Learned parameters $\theta$, $\{\alpha_i\}^M_{i=1}$, $\phi$. ($\alpha_0$ means no modulation)
\end{algorithmic}
\caption{The overall learning procedure}
\label{alg:main}
{\bfseries Step 1: Build a base network}
\begin{algorithmic}[1]
\STATE Build one large classification dataset $(x_{agg}, y_{agg})$ by aggregating all classes from $D_S$.
\STATE Learn $\theta$ by optimizing $f_e(x ; \theta, \alpha_{0})$ for the aggregated dataset ($\alpha_{0}$: no modulation).
\end{algorithmic}
{\bfseries Step 2: Add modulators through intra-domain episodic training}
\begin{algorithmic}[1]
\WHILE{not converged}
    \STATE Sample one domain $D_{S_i}$ from $D_S$, then sample one episode $(S, Q)$ from $D_{S_i}$.
    \STATE Learn $\alpha_i$ by optimizing $f_e(x; \theta, \alpha_{i})$ for $(S, Q)$ while keeping $\theta$ fixed.
\ENDWHILE
\end{algorithmic}
{\bfseries Step 3: Build a selection network through cross-domain episodic training} 
\begin{algorithmic}[1]
\WHILE{not converged}
    \STATE Sample one domain $D_{S_i}$ from $D_S$, then sample one episode $(S, Q)$ from $D_{S_i}$.
    \STATE Get a task representation $z_{task}$ by averaging embedding vectors of $S$ from the base network.
    \STATE Measure accuracies of $M+1$ models $\{f_e(x ; \theta, \alpha_i)\}^M_{i=0}$ for $(S, Q)$. ($\alpha_0$: no modulation.)
    \STATE Set the best model index $y_{sel}$ to the index of the model with the highest accuracy.
    \STATE Learn $\phi$ by training $f_s(z_{task} ; \phi)$ so as to predict $y_{sel}$ for $(S, Q)$.
\ENDWHILE
\end{algorithmic}
\end{algorithm}
\section{Additional experimental results}\label{sec:add_exp_results}

\subsection{5-way 1-shot classification}

\begin{figure*}[t!]
	\begin{subfigure}{\textwidth}
    	\centering
        \includegraphics[width=0.98\linewidth]{./figures/legends.png}
		\vspace{0.1in}
    \end{subfigure}
    \begin{subfigure}{0.48\textwidth}
    	\centering
        \includegraphics[width=\linewidth]{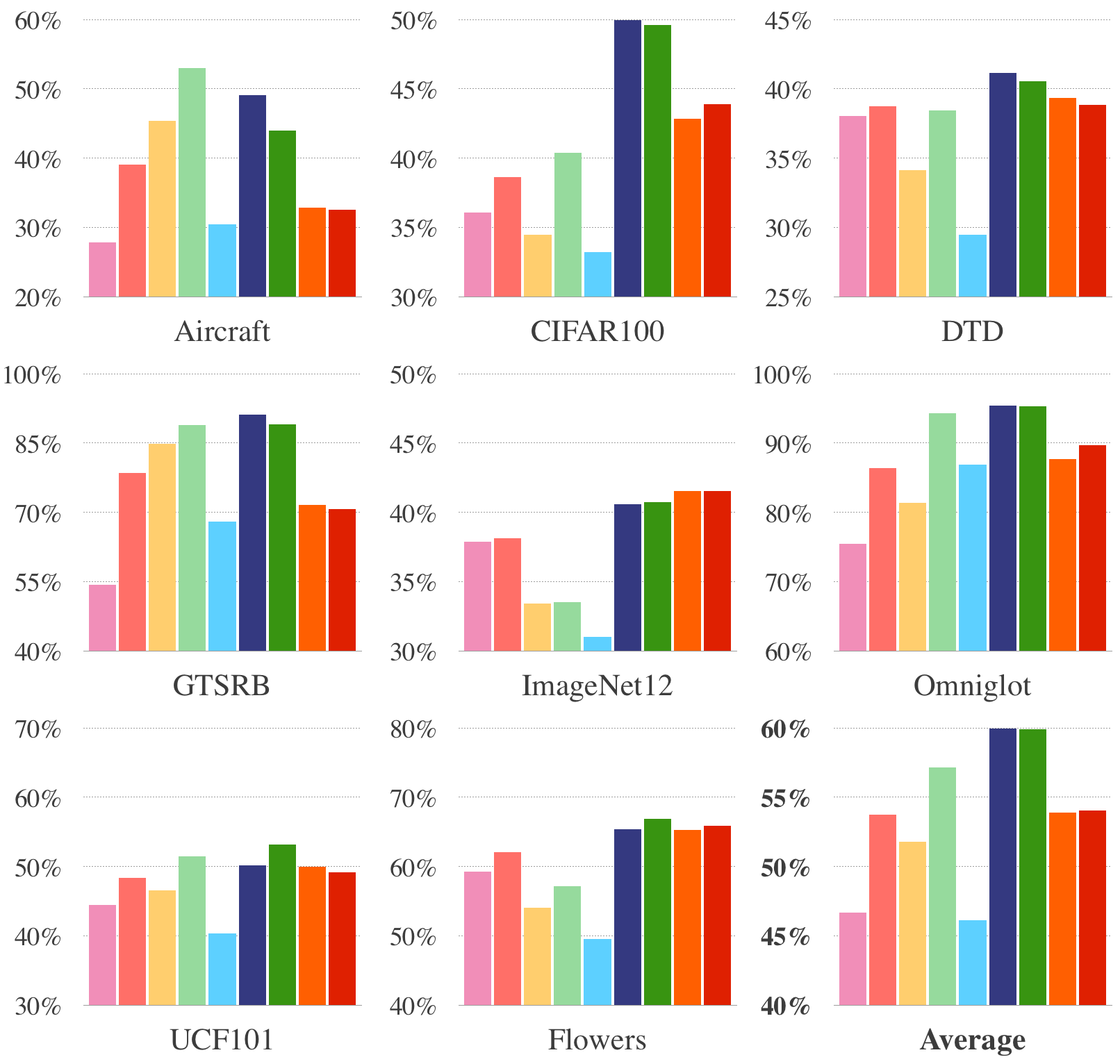}
		\caption{On seen domains.}
    \end{subfigure}
    \hfill
    \begin{subfigure}{0.48\textwidth}
        \centering
		\includegraphics[width=\linewidth]{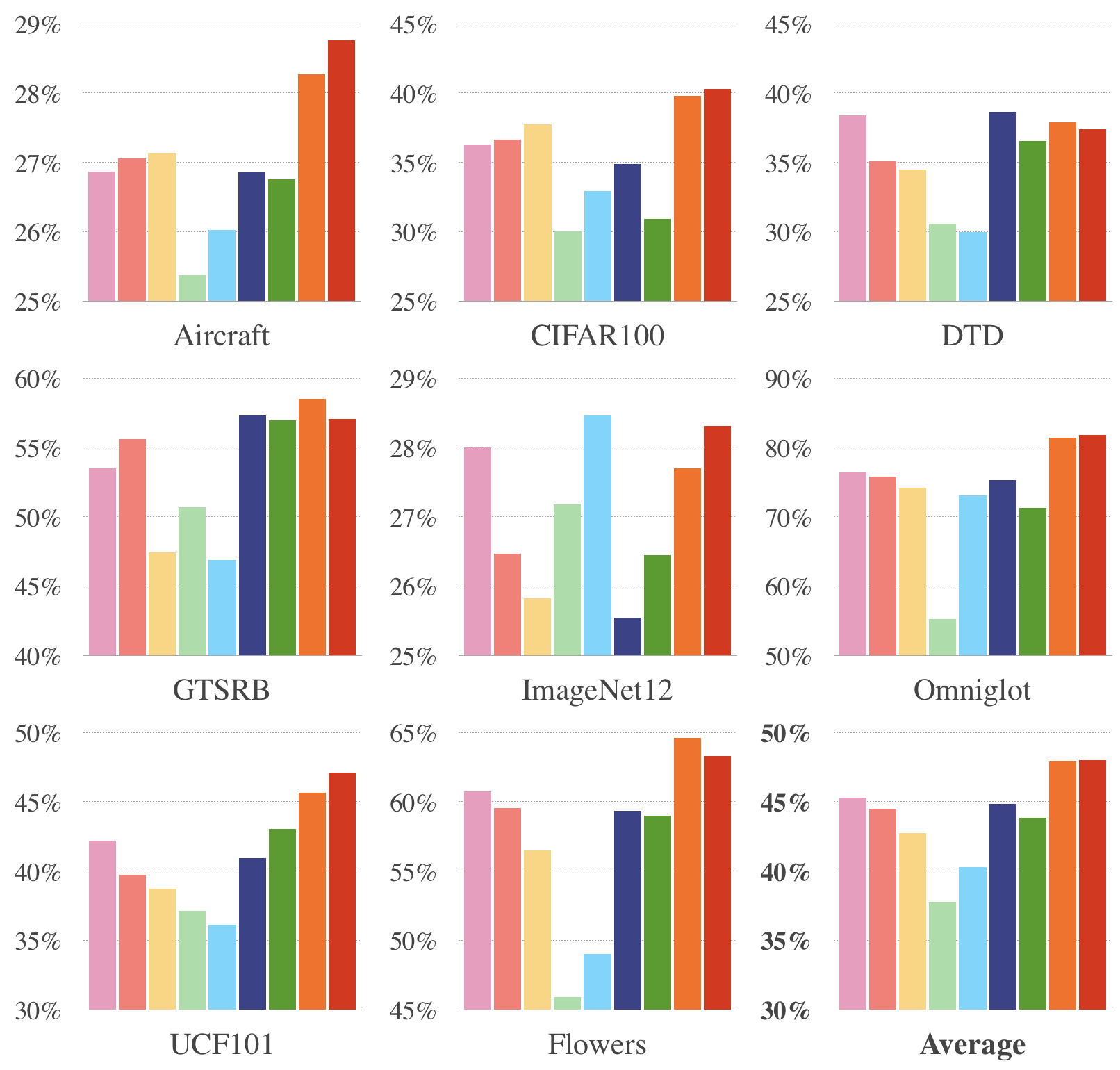}
		\caption{On unseen domains.}
    \end{subfigure}
    \caption{5-way 1-shot test accuracy on various datasets.
    }
    \vspace{-0.2cm}
    \label{figure:results_5w1s}
    \vspace{-0.2cm}
\end{figure*}

Figure \ref{figure:results_5w1s} shows test accuracy values of 5-way 1-shot classification tasks.

\subsection{Results without further adaptation}
We present the experimental results when we do not apply the further adaptation scheme introduced in \citep{chen2018a_closerLook}.
Specifically, \textit{ProtoNet}, \textit{FEAT}, and our models are tested without additional linear classifiers $f_c(\cdot; \psi)$. The number of parameter update steps in \textit{ProtoMAML} is reduced to 3, which is not enough to have the models fine-tuned.
Tables \ref{table:seen55_noFinetune} and \ref{table:unseen55_noFinetune} show the results tested on seen and unseen domains, respectively. We can see that accuracy drops in almost all cases compared to corresponding cases with further adaptation whose results are in Section \ref{sec:exp} in the main text, but our models generally do better than other methods in any experimental settings.


\begin{table*}[ht]
\fontsize{7}{7}\selectfont
    \caption{5-way 5-shot classification accuracy on seen domains without further adaptation.}
    \label{table:seen55_noFinetune}
    \vskip 0.15in
    \begin{center}
    \begin{sc}
    \begin{tabular}{ cccccccccc}
        \toprule
            &   \multicolumn{9}{c}{Methods} \\
            \cmidrule(r){2-10}
        Target
        & F-T  & Proto  & FEAT      & PMAML & S-Avg & DoS & DoS-Ch & DoA & DoA-Ch\\
        \midrule
        A
        & $39.53\%$ & $49.61\%$ & $58.53\%$ & $\mathbf{66.39\%}$ & $37.04\%$ & $62.86\%$ & $59.39\%$ & $42.82\%$ & $42.91\%$  \\
        C
        & $53.94\%$ & $57.48\%$ & $53.85\%$ & $59.37\%$ & $48.60\%$ & $69.94\%$ & $\mathbf{71.57\%}$ & $62.96\%$ & $64.37\%$ \\
        D
        & $55.68\%$ & $53.51\%$ & $52.91\%$ & $50.81\%$ & $42.98\%$ & $\mathbf{58.65\%}$ & $56.50\%$ & $58.43\%$ & $56.54\%$ \\
        G
        & $78.67\%$ & $86.21\%$ & $92.74\%$ & $95.22\%$ & $74.39\%$ & $96.49\%$ & $\mathbf{96.63\%}$ & $82.93\%$ & $83.31\%$ \\
        I
        & $54.64\%$ & $52.23\%$ & $49.91\%$ & $46.80\%$ & $41.69\%$ & $55.73\%$ & $55.49\%$ & $\mathbf{57.83\%}$ & $57.42\%$ \\
        O
        & $92.96\%$ & $96.20\%$ & $93.18\%$ & $98.98\%$ & $94.44\%$ & $98.86\%$ & $\mathbf{98.80\%}$ & $95.54\%$ & $96.70\%$ \\
        U
        & $62.78\%$ & $66.31\%$ & $64.76\%$ & $66.17\%$ & $59.15\%$ & $\mathbf{69.78\%}$ & $69.45\%$ & $69.68\%$ & $68.93\%$ \\
        F
        & $79.92\%$ & $83.46\%$ & $74.80\%$ & $75.84\%$ & $69.98\%$ & $83.60\%$ & $\mathbf{84.12\%}$ & $83.63\%$ & $83.28\%$ \\
        \midrule
        Average
        & $64.76\%$ & $68.13\%$ & $67.58\%$ & $69.95\%$ & $58.53\%$ & $\mathbf{74.49\%}$ & $73.99\%$ & $69.23\%$ & $69.18\%$ \\
        \bottomrule
    \end{tabular}
    \end{sc}
    \\ \raggedbottom{\footnotesize{A: Aircraft, C: CIFAR100, G: GTSRB, I: ImageNet12, O:Omniglot, U:UCF101, F:VGG-Flowers}}
    \\ \raggedbottom{\footnotesize{F-T: Fine-tune, Proto: ProtoNet, PMAML: ProtoMAML, S-Avg: Simple-Avg}}
    \end{center}
\end{table*}

\begin{table*}[ht]
\fontsize{7}{7}\selectfont
    \caption{5-way 5-shot classification accuracy on unseen domains without further adaptation.}
    \label{table:unseen55_noFinetune}
    \vskip 0.15in
    \begin{center}
    \begin{sc}
    \begin{tabular}{ cccccccccc}
        \toprule
            &   \multicolumn{9}{c}{Methods} \\
            \cmidrule(r){2-10}
        Target
            & F-T  & Proto  & FEAT      & PMAML & S-Avg & DoS & DoS-Ch & DoA & DoA-Ch\\
        \midrule
        A
            & $36.40\%$ & $35.42\%$ & $33.30\%$ & $32.87\%$ & $33.48\%$ & $34.73\%$ & $35.63\%$ & $36.93\%$ & $\mathbf{37.28\%}$ \\
        C
            & $53.26\%$ & $55.17\%$ & $52.60\%$ & $58.01\%$ & $49.23\%$ & $55.79\%$ & $50.09\%$ & $58.30\%$ & $\mathbf{59.22\%}$ \\
        D
            & $\mathbf{55.18\%}$ & $51.03\%$ & $50.37\%$ & $45.46\%$ & $45.99\%$ & $54.11\%$ & $48.81\%$ & $54.88\%$ & $54.37\%$ \\
        G
            & $78.01\%$ & $76.33\%$ & $75.79\%$ & $\mathbf{78.81\%}$ & $69.49\%$ & $77.03\%$ & $77.18\%$ & $77.36\%$ & $77.31\%$ \\
        I
            & $34.81\%$ & $32.90\%$ & $33.50\%$ & $36.46\%$ & $\mathbf{37.79\%}$ & $34.34\%$ & $32.73\%$ & $34.36\%$ & $35.27\%$ \\
        O
            & $92.69\%$ & $92.64\%$ & $91.99\%$ & $83.80\%$ & $91.45\%$ & $93.68\%$ & $91.99\%$ & $\mathbf{94.95\%}$ & $94.84\%$ \\
        U
            & $62.16\%$ & $59.74\%$ & $58.54\%$ & $58.40\%$ & $59.06\%$ & $62.04\%$ & $62.21\%$ & $65.93\%$ & $\mathbf{67.25\%}$ \\
        F
            & $81.00\%$ & $79.42\%$ & $80.82\%$ & $69.47\%$ & $75.39\%$ & $81.85\%$ & $82.69\%$ & $82.71\%$ & $\mathbf{83.89\%}$ \\
        \midrule
        Average
            & $61.69\%$ & $60.33\%$ & $59.62\%$ & $57.91\%$ & $57.74\%$ & $61.70\%$ & $60.17\%$ & $63.19\%$ & $\mathbf{63.68\%}$ \\
        \bottomrule
    \end{tabular}
    \end{sc}
    \\ \raggedbottom{\footnotesize{A: Aircraft, C: CIFAR100, G: GTSRB, I: ImageNet12, O:Omniglot, U:UCF101, F:VGG-Flowers}}
    \\ \raggedbottom{\footnotesize{F-T: Fine-tune, Proto: ProtoNet, PMAML: ProtoMAML, S-Avg: Simple-Avg}}
    \end{center}
\end{table*}

\subsection{Few-shot classification on varying number of source domains}

We conduct experiments with varying number of source datasets. 
Following the common real-world situation, we add from the largest dataset to the smallest one to our sources for meta-training. 
Tables \ref{table:246sources_seen} and \ref{table:246sources_unseen} show the experimental results with 2, 4, and 6 source datasets on seen and unseen domains respectively. 

Our selection and averaging methods outperform others consistently on seen and unseen domains similarly to the previous results.
Apart from comparing between the algorithms, it is commonly observed over all algorithms that the added source often harms the performance. 
For example, the CIFAR100 tasks tend to work poorer as the number of source datasets increases in the seen domain case.
This means that we should pay more attention to avoiding negative transfer between heterogeneous domains.

\begin{table*}[h]
\fontsize{7}{7}\selectfont
  \caption{Few-shot classification accuracy of varying number of sources on seen target domains.
   }
   
  \label{table:246sources_seen}
  \begin{center}
  \begin{sc}
  \begin{tabular}{cccccccccc}
     \toprule
            &&   \multicolumn{8}{c}{Methods} \\
            \cmidrule(r){3-10}
     S & T
        & Fine-tune & ProtoNet & FEAT & ProtoMAML & DoS & DoS-Ch & DoA & DoA-Ch\\
     \midrule
    \multirow{2}{*}{C,I}
            & C 
                & $54.72\%$ & $65.47\%$ & $65.39\%$ & $69.51\%$ & $\mathbf{73.04\%}$ & $72.63\%$ & $71.19\%$ & $69.58\%$\\
            & I 
                & $57.50\%$ & $55.88\%$ & $52.53\%$ & $56.37\%$ & $57.71\%$ & $57.84\%$ & $\mathbf{59.39\%}$ & $58.05\%$\\
    \cmidrule{1-10}
    \multicolumn{2}{c}{Average}
                & $56.11\%$ & $60.68\%$ & $58.96\%$ & $62.94\%$ & $\mathbf{65.38\%}$ & $65.24\%$ & $65.29\%$ & $63.82\%$\\
    \cmidrule{1-10}
    \multirow{4}{*}{C,G,I,O}
            & C 
                & $54.24\%$ & $58.31\%$ & $63.82\%$ & $68.93\%$ & $\mathbf{71.11\%}$ & $70.15\%$ & $68.14\%$ & $66.67\%$\\
            & G 
                & $81.21\%$ & $90.58\%$ & $95.40\%$ & $96.61\%$ & $\mathbf{96.96\%}$ & $96.36\%$ & $92.94\%$ & $91.55\%$\\
            & I 
                & $55.70\%$ & $52.04\%$ & $50.58\%$ & $53.72\%$ & $55.78\%$ & $56.10\%$ & $\mathbf{58.04\%}$ & $57.74\%$\\
            & O 
                & $94.82\%$ & $97.14\%$ & $95.10\%$ & $\mathbf{99.52\%}$ & $98.94\%$ & $98.73\%$ & $97.86\%$ & $97.64\%$\\
    \cmidrule{1-10}
    \multicolumn{2}{c}{Average}
                & $71.49\%$ & $74.52\%$ & $76.23\%$ & $79.70\%$ & $\mathbf{80.70\%}$ & $80.33\%$ & $79.25\%$ & $78.40\%$\\
    \cmidrule{1-10}
            & A 
                & $40.68\%$ & $51.92\%$ & $62.77\%$ & $\mathbf{70.78}\%$ & $61.94\%$ & $57.49\%$ & $47.77\%$ & $46.86\%$\\
            & C 
                & $54.03\%$ & $58.68\%$ & $56.56\%$ & $60.47\%$ & $69.54\%$ & $\mathbf{70.32\%}$ & $66.48\%$ & $65.41\%$\\
    A,C,G,  & G 
                & $76.35\%$ & $89.95\%$ & $95.64\%$ & $96.50\%$ & $\mathbf{97.71\%}$ & $96.78\%$ & $89.65\%$ & $89.62\%$\\
    I,O,U   & I 
                & $53.37\%$ & $52.80\%$ & $48.62\%$ & $51.74\%$ & $54.59\%$ & $56.03\%$ & $57.34\%$ & $\mathbf{58.03\%}$\\
            & O 
                & $94.06\%$ & $96.94\%$ & $95.30\%$ & $\mathbf{99.11\%}$ & $98.89\%$ & $98.76\%$ & $97.20\%$ & $97.92\%$\\
            & U 
                & $62.01\%$ & $65.49\%$ & $62.20\%$ & $67.11\%$ & $70.44\%$ & $70.23\%$ & $69.16\%$ & $\mathbf{71.42\%}$\\
    \cmidrule{1-10}
    \multicolumn{2}{c}{Average}
                & $63.42\%$ & $69.30\%$ & $70.18\%$ & $74.29\%$ & $\mathbf{75.52\%}$ & $74.94\%$ & $71.27\%$ & $71.54\%$\\
     \bottomrule
  \end{tabular}
  \end{sc}
  \\ \raggedbottom{\footnotesize{S: source datasets, T: target dataset}}
  \\ \raggedbottom{\footnotesize{A: Aircraft, C: CIFAR100, G: GTSRB, I: ImageNet12, O: Omniglot, U: UCF101, F: Flowers}}
  \end{center}
\end{table*}

\begin{table*}[h]
\fontsize{7}{8}\selectfont
  \caption{Few-shot classification accuracy of varying number of sources on unseen target domains.
  }
  \label{table:246sources_unseen}
  \begin{center}
  \begin{sc}
  \begin{tabular}{ cccccccccc}
    \toprule
        &&   \multicolumn{8}{c}{Methods} \\
    \cmidrule(r){3-10}
    S   & T
        & Fine-tune  & ProtoNet  & FEAT      & ProtoMAML & DoS & Dos-Ch & DoA & DoA-ch\\
    \midrule
        \multirow{6}{*}{C,I}
        &A&
        $38.65\%$ & $39.38\%$ & $36.71\%$ & $35.37\%$ & $38.43\%$ & $39.27\%$ & $39.14\%$ & $\mathbf{40.57\%}$ \\
        &D&
        $53.89\%$ & $55.27\%$ & $52.39\%$ & $51.04\%$ & $54.80\%$ & $54.90\%$ & $\mathbf{57.15\%}$ & $56.99\%$ \\
        &G&
        $70.01\%$ & $81.37\%$ & $77.67\%$ & $81.25\%$ & $\mathbf{83.83\%}$ & $82.65\%$ & $80.13\%$ & $79.93\%$ \\
        &O&
        $92.72\%$ & $92.62\%$ & $90.61\%$ & $91.37\%$ & $93.11\%$ & $93.31\%$ & $94.00\%$ & $\mathbf{94.34}\%$ \\
        &U&
        $63.80\%$ & $63.25\%$ & $60.04\%$ & $59.89\%$ & $63.67\%$ & $64.81\%$ & $66.28\%$ & $\mathbf{67.15\%}$ \\
        &F&
        $79.28\%$ & $81.77\%$ & $78.86\%$ & $80.50\%$ & $81.26\%$ & $80.84\%$ & $\mathbf{82.96\%}$ & $82.85\%$ \\
    \cmidrule{1-10}
    \multicolumn{2}{c}{Average}
        & $66.39\%$ & $68.94\%$ & $66.05\%$ & $66.63\%$ & $69.18\%$ & $69.30\%$ & $69.94\%$ & $\mathbf{70.31\%}$\\
    \cmidrule{1-10}
        \multirow{4}{*}{C,G,I,O}&
        A
        & $38.12\%$ & $36.88\%$ & $35.11\%$ & $34.96\%$ & $37.90\%$ & $38.04\%$ & $39.66\%$ & $\mathbf{39.99\%}$ \\
        &D
        & $55.04\%$ & $49.96\%$ & $49.55\%$ & $49.79\%$ & $55.85\%$ & $56.53\%$ & $\mathbf{56.73\%}$ & $55.67\%$ \\
        &U
        & $63.25\%$ & $56.53\%$ & $60.93\%$ & $64.12\%$ & $64.32\%$ & $64.84\%$ & $67.58\%$ & $\mathbf{67.69\%}$ \\
        &F
        & $79.88\%$ & $76.79\%$ & $77.67\%$ & $81.40\%$ & $80.69\%$ & $79.54\%$ & $\mathbf{83.25\%}$ & $82.20\%$ \\
    \cmidrule{1-10}
    \multicolumn{2}{c}{Average}
        & $59.07\%$ & $55.04\%$ & $55.82\%$ & $57.57\%$ & $59.69\%$ & $59.74\%$ & $\mathbf{61.81\%}$ & $61.39\%$\\
    \cmidrule{1-10}
        A,C,G
        &D
        & $53.60\%$ & $51.52\%$ & $51.86\%$ & $50.94\%$ & $55.54\%$ & $55.69\%$ & $56.71\%$ & $\mathbf{57.07\%}$ \\
        I,O,U
        &F
        & $80.80\%$ & $80.58\%$ & $79.70\%$ & $79.40\%$ & $81.62\%$ & $81.66\%$ & $84.30\%$ & $\mathbf{84.50\%}$ \\
    \cmidrule{1-10}
    \multicolumn{2}{c}{Average}
        & $67.20\%$ & $66.05\%$ & $65.78\%$ & $65.17\%$ & $68.58\%$ & $68.68\%$ & $70.51\%$ & $\mathbf{70.79\%}$ \\
     \bottomrule
  \end{tabular}
  \end{sc}
  \\ \raggedbottom{\footnotesize{S: source datasets, T: target dataset}}
  \\ \raggedbottom{\footnotesize{A: Aircraft, C: CIFAR100, G: GTSRB, I: ImageNet12, O: Omniglot, U: UCF101, F: Flowers}}
  \end{center}
\end{table*}

\subsection{Comparative analysis about the averaging methods}

\begin{figure}
\centering
    \begin{subfigure}{\textwidth}
        \includegraphics[width=1.0\linewidth]{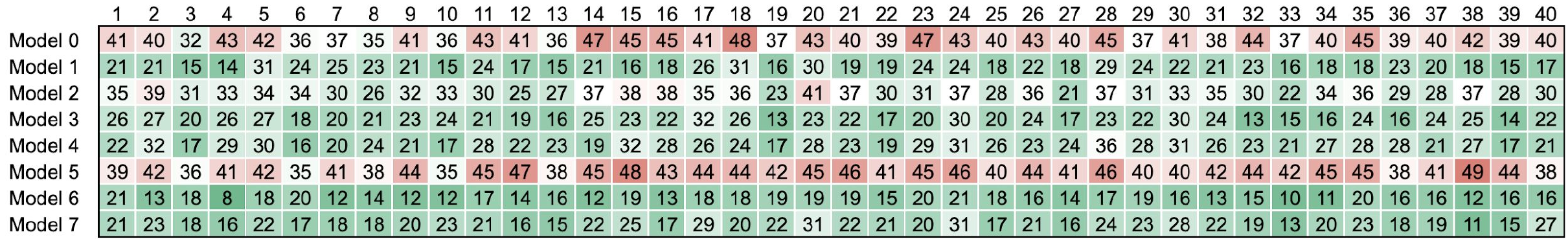}
        \caption{Simple averaging (Simple-Avg)}
    \end{subfigure}
    \begin{subfigure}{\textwidth}
        \includegraphics[width=1.0\linewidth]{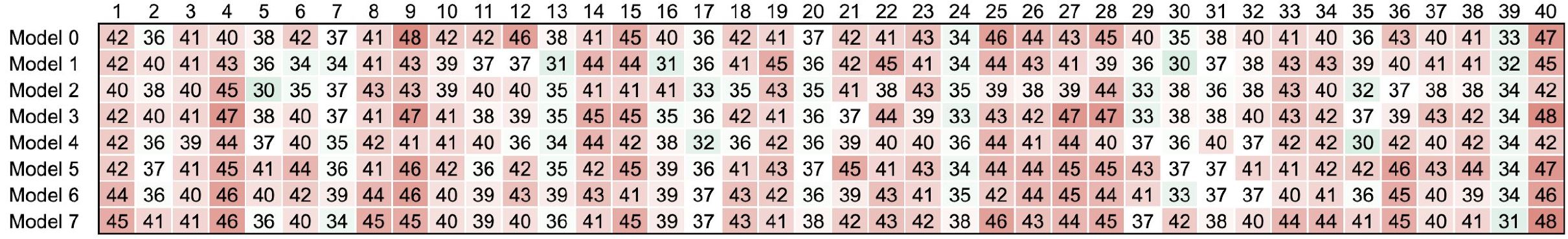}
        \caption{Proposed averaging (DoA)}
    \end{subfigure}
\caption{Contributions of individual models in model averaging methods.}   
\label{fig:comp_ensemble}
\end{figure}

As an effort for better understanding the averaging methods, we investigate how each model in the pool contributes to the final prediction. 
Figures \ref{fig:comp_ensemble}(a) and \ref{fig:comp_ensemble}(b) show how many correct predictions are made by each model with the \textit{Simple-Avg} and our \textit{DoA} methods respectively given 50 queries per episode for 40 episodes. 

The measured numbers show that the individual models of our \textit{DoA} perform better than those in the \textit{Simple-Avg}, which explains the higher performance of the proposed method partly. 
Additionally, we can observe that major contributors (i.e., the models with higher accuracy) tend to change every episode in our \textit{DoA} whereas only two models seem to play dominant roles regardless of the given episode. 
This implies that our method for constructing the model pool provides the averaging model with more beneficial diversity.

\section{The loss function for the selection network} 

Equation (\ref{eq:learn_to_select}) shows the loss ($loss_{sel}$) used to train the selection network $f_s(\cdot; \phi)$. Here, $acc_i$ is the classification accuracy of the model with the modulator parameterized by $\alpha_i$ in the pool.
The accuracy is measured for query examples in a given episode by making a prediction in the same way with the ProtoNet \cite{DBLP:conf/nips/SnellSZ17_protoNet}, where the class whose prototype is the closest to the embedding vector of a given query example is picked as the final prediction.

\begin{equation}
    \begin{aligned}
        & z_{task} = \frac{1}{NK}\sum^{NK}_{i=1}f_e(x^s_i; \theta, \alpha_0)\\
        & \hat{y}_{sel} = \textnormal{softmax}(f_s(z_{task}; \phi)  )\\
        & y_{sel} = \textnormal{argmax}(\{acc_i(\{x^s, y^s\}^{NK}_{i=1}, \{x^q, y^q\}^T_{j=1})\}^M_{i=0})\\
        & loss_{sel} = \textnormal{cross\_entropy}(\hat{y}_{sel}, y_{sel})
    \end{aligned}
    \label{eq:learn_to_select}    
\end{equation}



\end{document}